\documentclass[letterpaper, 10 pt, conference]{ieeeconf}  

\usepackage{tabularx,booktabs}
\usepackage[flushleft]{threeparttable}
\usepackage{nameref}
\usepackage{amsmath}
\usepackage{pgfplots}
\pgfplotsset{compat=1.18}
\usepackage[utf8]{inputenc}
\usepackage{booktabs}
\usepackage{cite}
\usepackage{graphicx}
\usepackage{color}
\usepackage{url}
\usepackage{cite}
\usepackage{comment}
\usepackage{amsfonts}
\usepackage[per-mode=symbol]{siunitx}
\usepackage{mathtools}
\usepackage{setspace}
\usepackage{bm}
\usepackage{algorithm,algorithmic}

\let\labelindent\relax
\usepackage{enumitem}

\IEEEoverridecommandlockouts                              
\title{\LARGE \bf A Biomechanics-Inspired Approach to Soccer Kicking for \\Humanoid Robots}

\author{Daniel Marew$^{1}$, Nisal Perera$^{1}$, Shangqun Yu$^{1}$, Sarah Roelker$^{2}$, and Donghyun Kim$^{1}$
\thanks{$^{1}$ College of Information and Computer Sciences, University of Massachusetts Amherst, MA, U.S. ({\tt\small donghyunkim@cs.umass.edu}) }
\thanks{$^{2}$ School of Public Health and Health Sciences, University of Massachusetts Amherst, MA, U.S.}
}

\begin{document}

\maketitle
\thispagestyle{empty}
\pagestyle{empty}

\begin{abstract}
Soccer kicking is a complex whole-body motion that requires intricate coordination of various motor actions. To accomplish such dynamic motion in a humanoid robot, the robot needs to simultaneously: 1) transfer high kinetic energy to the kicking leg, 2) maintain balance and stability of the entire body, and 3) manage the impact disturbance from the ball during the kicking moment. Prior studies on robotic soccer kicking often prioritized stability, leading to overly conservative quasi-static motions. In this work, we present a biomechanics-inspired control framework that leverages trajectory optimization and imitation learning to facilitate highly dynamic soccer kicks in humanoid robots. We conducted an in-depth analysis of human soccer kick biomechanics to identify key motion constraints. Based on this understanding, we designed kinodynamically feasible trajectories that are then used as a reference in imitation learning to develop a robust feedback control policy. We demonstrate the effectiveness of our approach through a simulation of an anthropomorphic 25 DoF bipedal humanoid robot, named PresToe, which is equipped with 7 DoF legs, including a unique actuated toe. Using our framework, PresToe can execute dynamic instep kicks, propelling the ball at speeds exceeding $11~\si{\meter\per\second}$ in full dynamics simulation. 
\end{abstract}

\section{Introduction}
The RoboCup Federation has established an ambitious goal for 2050: to develop a humanoid soccer team capable of defeating the human FIFA World Cup champions of that year~\cite{Kitano1998, Gerndt2015}. For this dream to become a reality and to effectively compete with humans, the robots must not only walk, run, and leap, but also master the art of kicking a ball as swiftly and as accurately as humans. Achieving a powerful and accurate kick in bipedal robots is a complex endeavor. It requires the substantial transfer of kinetic energy to the kicking leg, all while maintaining balance with only one foot grounded, and ensuring effective recovery after ball impact. However, existing studies on robotic soccer kicks have not addressed the full spectrum of the challenges associated with such dynamic motion. For example, \cite{cmu, Haddadin2009, Vahidi2015} anchored the robot to the ground, thereby circumventing the challenges of balance and post-kick recovery. While this allows researchers to concentrate on the dynamic interaction between the foot and ball during impact, the results are hard to extend to biped robots with significant under-actuation. In contrast, \cite{nao, barrett2010controlled, Hester2010, Leottau2015, sung2011} explore soccer kick motions using fully mobile humanoid robots. However, these approaches tend to favor conservative stationary kick movements, resulting in less powerful kicks to guarantee stability both before and after ball contact.

One common missing perspective in both approaches is the lack of focus on momentum-building preparatory steps before a kick. In human soccer, players use such steps to accumulate momentum, enabling more powerful strikes. Many robotic approaches overlook this critical aspect, often keeping the robot's supporting leg stationary, which limits the kinetic energy transferred to the ball at impact. There exists only a handful of research utilizing stepping before kicking~\cite{Teixeira2020, Rezaeipanah2020, asimo_demo_2014}. However, \cite{Teixeira2020, Rezaeipanah2020} used the walking step mainly to reposition the robot rather than to build momentum for a stronger kick. A video demonstration by Honda Research featuring the Asimo robot \cite{asimo_demo_2014} shows more powerful kicks achieved in part through preparatory steps. However, the robot's motion lacks key human-like traits of an instep kick: it omits arm and torso movements crucial for angular momentum regulation and balance, and follows a straight-line trajectory instead of a human's typical curved path that is associated with more powerful kicks \cite{Shan2011}. Consequently, the kick resembles a walk-and-kick motion rather than a dynamic human instep kick.

\begin{figure}
    \centering
    \includegraphics[width=\linewidth]{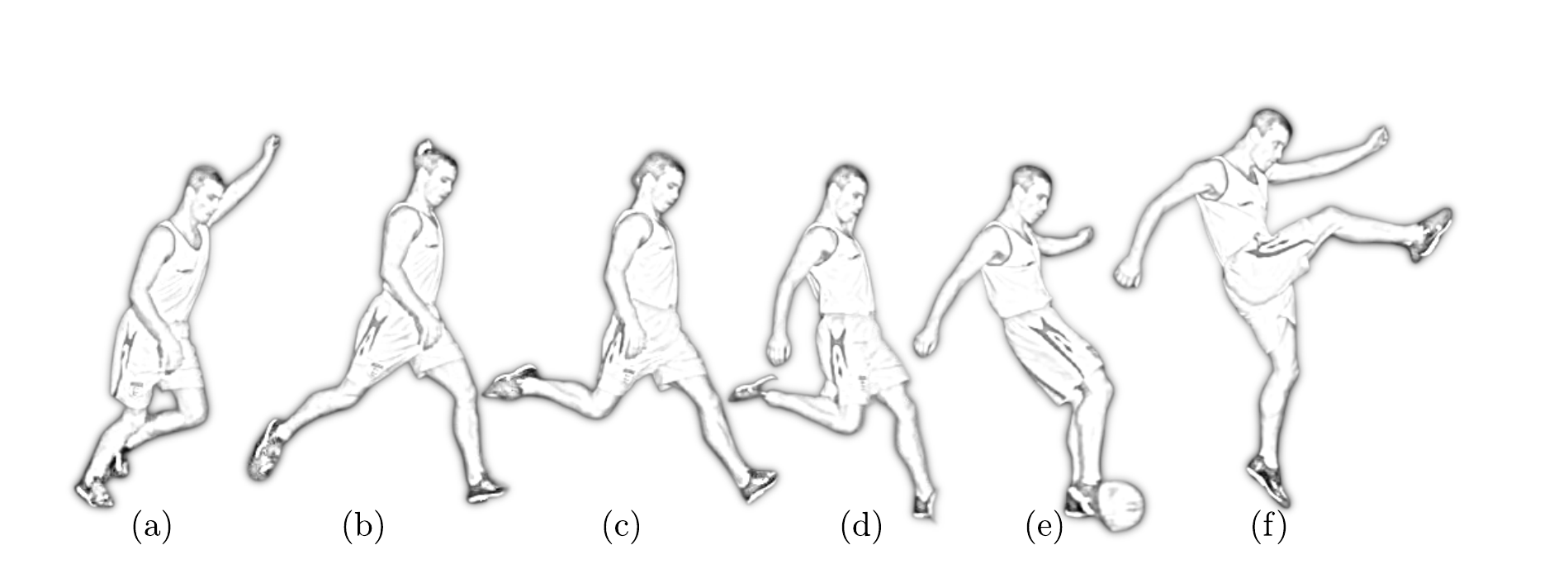}
    \caption{Phases of an in-step soccer kick: (a) Approach/Run-up, (b) Planting/Support, (c) Wind-up/Backswing, (d) Cocking, (e) Swing/Acceleration, and (f) Follow-through. Adapted from \cite{bio_youtube_video}.}
    \label{fig:instep-kick}
    \vspace{-3mm}
\end{figure}

Often, Biomechanics provides interesting perspectives and useful information for robot control. The previous studies of soccer kick biomechanics identified several key attributes that determine the power and quality of a kick \cite{Shan2011,Shan2022}. Among these, the foremost determinant of kick power is the kinodynamic whip action~\cite{Shan2011, Shan2022}. This action results from the initial loading of the trunk, hip, and knee joints, often referred to as the tension arc \cite{Shan2011, Shan2022}. Potential energy stored in the hip flexors, quadriceps, hamstrings, and calf muscles is then quickly released in a coordinated manner, moving from proximal to distal parts. This sequential action, resembling a whip, transfers large kinetic energy to the kicking foot and subsequently to the ball. 

Another key attribute is the approach angle towards the ball, which for professional players typically ranges between $24~\si{\degree}$ and $43~\si{\degree}$~\cite{Egan2007, Shan2011, Shan2022}. This approach angle enhances kicking power by permitting a greater range of motion for trunk rotation, thus leading to a more substantial loading of the trunk prior to the whip action's onset \cite{Shan2022}.
Furthermore, an equally pivotal attribute is the placement of the supporting foot. In this regard, expert advice emphasizes positioning the foot next to and slightly behind the ball, parallel to the desired kick direction. Moreover, the length of the step taken immediately before planting the foot strongly correlates with kick power, with longer strides resulting in more powerful kicks \cite{Shan2011, Shan2022}.
Lastly, the ankle locks before impact to ensure a firmer surface contact, which, in turn, correlates with greater power transfer \cite{barfield1998biomechanics}.

In this work, we focus on the instep soccer kick which is, a powerful and widely studied skill, comprises six stages as illustrated in Fig.~\ref{fig:instep-kick}: (a) the approach or run-up, where the player advances towards the ball; (b) the planting or support phase, involving placement of the non-kicking foot; (c) the wind-up or backswing, where the kicking leg is retracted; (d) the cocking phase, characterized by forward thigh motion with a flexed knee; (e) the swing or acceleration phase, involving rapid hip flexion and knee extension; and (f) the follow-through, continuing the leg's forward motion after ball impact \cite{barfield1998biomechanics, Lees1998, Kellis2007, Brophy2007, Lees2010, Shan2011, Shan2022}.
\begin{figure}
    \centering
    \includegraphics[width=\linewidth]{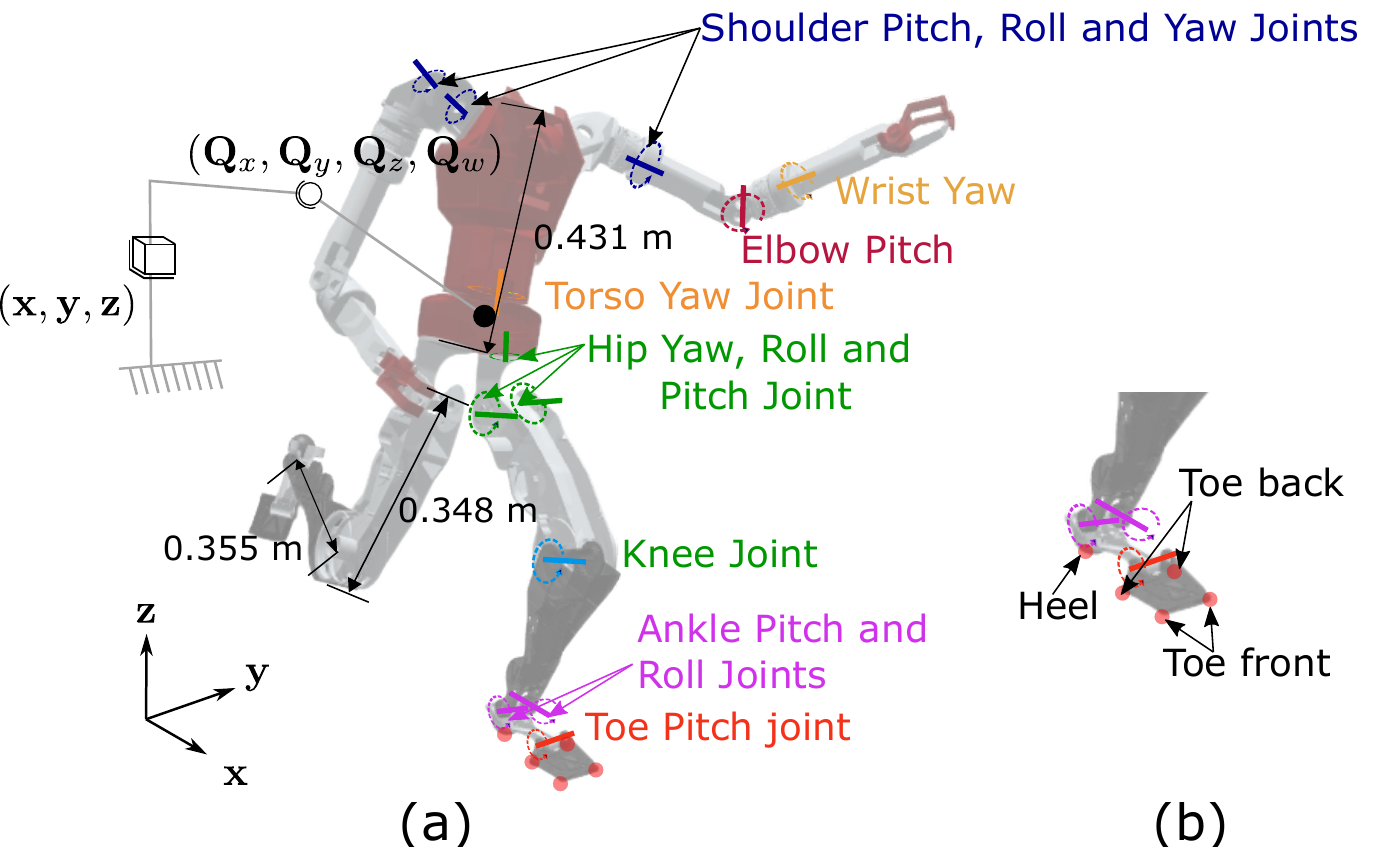}
    \caption{PresToe: (a) Robot degrees of freedom (b) Foot contact points}
    \label{fig:prestoe}
\end{figure}


Our strategy to implement the principles learned from biomechanics into robot control is developing a novel motion planning and control framework that combines motion retargeting, kino-dynamic trajectory optimization, and imitation learning. Our framework is divided into two parts: 1) motion planning based on human motion capture data, and 2) imitation learning to find robust control policy using the found motions as reference. In the motion planning phase, we use kinodynamic trajectory optimization \cite{dai} to properly account for the system's physical limits, which are often ignored in human-to-robot motion retargeting studies~\cite{tang2023humanmimic, yan2023imitationnet, cheng2024expressive, he2024learning}. Considering the robot's dynamics is crucial not only because the resultant motions are within the joint torque/velocity/position limits, but also due to the balance stability. \cite{nakaoka2005task} demonstrated effective retargeting of human dance motion by using a Linear Inverted Pendulum Model (LIPM) of the robot and enforcing Zero Moment Point (ZMP) constraints. Furthermore, \cite{rempe2020contact} showed that incorporating dynamics leads to more realistic-looking retargeting of human motion to animated characters. Their approach uses a two-stage process: first, obtaining a trajectory based on a Single Rigid Body (SRB) model, then performing inverse kinematics to recover a kinematic trajectory consistent with the Center of Mass (CoM) and foot contact locations. 

Our approach is comprehensive, using kinodynamic trajectory optimization that leverages the centroidal dynamics of the robot, which are more expressive than SRB or LIPM approximations. Additionally, our resulting kinematic trajectory aligns not only with the CoM and foot contact positions but also with linear and angular momenta, yielding realistic and expressive motion retargeting. Beyond modeling and optimization, we incorporate biomechanical observations from the six phases of the instep kick as constraints and cost terms in our trajectory optimization formulation, complementing the implicit priors obtained from the MoCap data.

After obtaining optimal kicking trajectories that include essential preparatory momentum-building steps, we train a reinforcement learning control policy with an imitation objective. This policy aims to track the trajectories obtained in the motion planning phase while maintaining the robot's balance in Isaac Gym \cite{makoviychuk2021isaac} -- accounting for factors such as the dynamic interaction between the robot and ball.

The main contributions of this paper are as follows:
1) We present a new motion planning and imitation learning frameworks for performing human-like soccer kicks by identifying key constraints and insights from the biomechanics of soccer kicks.
2) We showcase instep kicks that propel the ball at speeds exceeding $11~\si{\meter\per\second}$. This is almost double the 6.6 m/s instep kick achieved by the ground-anchored single-leg robot~\cite{Haddadin2009}. Notably, the demonstrated ball speed is about 40\% of the speed an average human player can achieve \cite{radja2019ball}, which is an impressive result considering the robot's size and limited torque capability. The robot weighs 30kgs and has a maximum hip pitch torque of only 48 N.m, compared to the average human player's hip joint flexion/extension capability of over 250 Nm \cite{sakamoto2016kinetic}.
3) We highlight the importance of proper consideration of dynamics in the offline motion planning stage for sample-efficient imitation learning. This is demonstrated by comparing purely kinematics-based reference motions with kinodynamically consistent ones.
\section{Methodology}
\begin{figure}
    \centering
    \includegraphics[width=\linewidth]{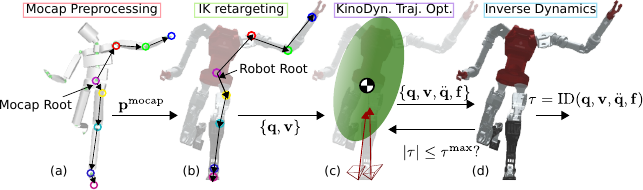}
\caption{Offline Motion Planning Steps: (a) MoCap Link Rescaling: Corresponding joints are highlighted in the same color (on the MoCap skeleton in (a) and robot in (b)). For clarity, only the left-side joint correspondence between the MoCap data and the robot is shown. (b) Inverse kinematics-based retargeting, ignoring dynamics. (c) Kinodynamic trajectory optimization. (d) Full-body dynamics-based torque limit violation verification.}
    \label{fig:offline-framework}
\end{figure}
    \subsection{Offline Kicking Trajectory Generation}
Our objective is to emulate the powerful kicks demonstrated by humans using a robotic platform. While earlier works attempted to manually craft kicking motions from scratch,  we believe in harnessing the natural expertise of humans \cite{nao, barrett2010controlled, Hester2010, Leottau2015, sung2011}. To this end, we utilize MoCap data derived from humans executing soccer kicks as a reference for generating robot trajectories that are consistent with the robot's kinematic and torque capability. Our methodology comprises four stages. Initially, we preprocess the MoCap data to ensure it aligns with the robot's morphology. Subsequently, using inverse kinematics, we derive purely kinematic joint position and velocity trajectories that closely mirror the link position trajectory present in the MoCap data. These kinematic trajectories serve both as a reference and an initial guess to inform the more complex, non-linear kinodynamic trajectory optimization problem. Here, we generate trajectories that are consistent with the robot's centroidal dynamics and kinematics. Finally, we use inverse dynamics to verify that the robot's torque limits are not violated, and we make necessary adjustments to the trajectory if any violations are detected. 
A detailed depiction of our framework is illustrated in Fig. \ref{fig:offline-framework}.
\label{offlineTO}
\subsubsection{Mocap Data Preprocessing}
 To derive a robot trajectory compatible with the robot's morphology, we preprocess the original MoCap data by adjusting link lengths to align with the robot's structure. We initiate this process by establishing a correspondence between human and robot joints, as shown in Fig. \ref{fig:offline-framework} (a). Starting from the root joint, we recursively scale the link lengths in the MoCap data to achieve a match.
Moreover, during this stage, we establish a ground foot contact schedule and other important timing parameters. Specifically, we identify the beginning of the swing phase ($T^{\text{swing}}$), the ankle locking time ($T^{\text{lock}}$), and the end of the ball impact phase ($T^{\text{impact}}$). We derive the ground contact schedule by monitoring when the feet initiate or break contact with the ground, using heuristics based on foot height and velocity thresholds \cite{rempe2020contact}. This information is vital for subsequent stages as it enables the imposition of constraints that prevent unwanted artifacts such as foot skating.
\subsubsection{Kinodynamic Trajectory Optimization}
\label{sec:kto}
We use kinodynamic trajectory optimization, first proposed by \cite{dai}, to generate physically consistent human-like soccer kicking trajectories, using the solution from the prior step as a reference. This approach, based on the robot's centroidal dynamics and full-body kinematics, strikes a balance between full-body dynamics and reduced-order models like Linear Inverted Pendulum (LIP) or Single Rigid Body (SRB) based approaches.
It has fewer non-linearities compared to full-body dynamics-based trajectory optimization, hence can be solved more efficiently while being more realistic and expressive than LIP or SRB models, which rely on restrictive assumptions \cite{dai, MIT_humanoid}.
In kinodynamic trajectory optimization, the centroidal dynamics of the robot, given by Eqs. \eqref{eq:cen_lindyn} and \eqref{eq:cen_angdyn}, relates the ground reaction forces to the robot's linear and angular momentum, while the kinematics model is used to determine a corresponding kinematic trajectory consistent with the center of mass, contact location, and momenta. 
Thus, the optimization encompasses centroidal variables such as the position $\mathbf{r} \in \mathbb{R}^3$, velocity $\mathbf{\dot{r}} \in \mathbb{R}^3$, and acceleration $\mathbf{\ddot{r}} \in \mathbb{R}^3$ of the robot's center of mass (CoM), the robot's centroidal angular momentum (CAM) $\mathbf{h} \in \mathbb{R}^3$ and its rate $\mathbf{\dot{h}} \in \mathbb{R}^3$, the location of contact points $\mathbf{c}_{i}\in \mathbb{R}^3$, as well as the external ground reaction forces acting at the contact points $\mathbf{f}_i \in \mathbb{R}^3$. The optimization also involves kinematic variables: joint position $\mathbf{q} \in \mathbb{R}^{n+7}$ and joint velocity $\mathbf{v} \in \mathbb{R}^{n+6}$, where $n$ is the number of joints of the robot. The additional seven dimensions in the position vector refer to the body position $\in \mathbb{R}^3$ and orientation represented by a unit quaternion $\in \mathbb{R}^4$, while the velocity vector includes six dimensions for the body's linear and angular velocities.
\begin{align}
\label{eq:cen_lindyn}
m\mathbf{\ddot{r}} &= \sum_{i=1}^{n_c} \mathbf{f}_{i} + m\mathbf{g} \\
\label{eq:cen_angdyn}
\mathbf{\dot{h}} &= \sum_{i=1}^{n_c} (\mathbf{c}_{i} - \mathbf{r}) \times \mathbf{f}_{i}
\end{align}
Here, $n_c$ represents the number of contact points. For the PresToe robot (see Fig. \ref{fig:prestoe}) in this study, we utilize 10 contact points: 5 for each foot. Specifically, there are two contact points on the front part of the toe, two on the back part, and one for the heel, as depicted in Fig. \ref{fig:prestoe}(b).
Following \cite{dai}, the continuous dynamics of the robot is sampled at regular intervals of $\Delta t$, in order to transcribe the optimization problem into a nonlinear programming problem. Equations \ref{eq:cen_lindyn}-\ref{eq:cen_angdyn} describe the discretized linear and angular centroidal dynamics of the robot, respectively at timestep $k$. The optimization objective comprises several components.
Firstly, reference tracking objective represented by     
\begin{equation}
    J_{k}^{\rm ref} =  
    \|\phi_{FK}(\mathbf{q}_k) - \mathbf{p}^{\text{mocap}}_k\|^2_{\mathbf{Q}_{m}}    
\end{equation}
which ensures the resulting trajectory is close to the one obtained in the previous step.
Additionally, we included a cost term to encourage kicking foot acceleration during the swing phase of the motion. This term is applied from the start of the swing phase to the end of the impact phase, corresponding to phases (d) and (e) in Fig. \ref{fig:instep-kick}.
\begin{equation}
J^{\rm imp}_k = - \zeta_k \|\bm{J}_{foot, k}(\mathbf{q}_k)\mathbf{v}_k\|^2_{\bm{Q}{i}}
\end{equation}
where $\zeta_k$ is a weighting function given by
\begin{equation}
\zeta_k =
\begin{cases}
1 & \text{if }  k\Delta t \in \left [T^{\text{swing}}, T^{\text{impact}}\right], \\
0 & \text{otherwise}.
\end{cases}
\end{equation}
Here $\bm{J}_{foot, k}$ is the Jacobian of the kicking leg's foot link in the world frame.
Lastly, a regularization cost is included as 
\begin{equation}
J_{k}
^{\rm reg} = (1-\zeta_k)\|\mathbf{v}_k\|^2_{Q_v} + \|\mathbf{h}_k\|^2_{Q_h} + \sum_i^{n_c}\|\mathbf{f}_{i,k}\|^2_{Q_f}
\end{equation}
Where ${Q_m}$, ${Q_v}$, ${Q_i}$, ${Q_h}$ and ${Q_f}$ positive definite diagonal weighting matrices. 
In addition to the dynamic constraints given by Eqs. \eqref{eq:cen_lindyn}-\eqref{eq:cen_angdyn}, we enforce consistency between the centroidal and kinematic variables using Eqs. \eqref{eq:ang_consistency} to \eqref{eq:contact_loc}.
\begin{align}
        \label{eq:ang_consistency}
        &\boldsymbol{h}_k = \mathbf{A}_{CAM}(\mathbf{q}_k)\mathbf{v}_k\\
        \label{eq:com_pos_consistency}
    &\boldsymbol{r}_k = \phi_{CoM}(\mathbf{q}_k)\\
    \label{eq:contact_loc}
    &\boldsymbol{c}_{i,k} = \phi_{c_{i}}(\mathbf{q}_k)
\end{align}
where $\mathbf{A}_{CAM}$ represents the centroidal angular momentum matrix \cite{dai}. In this context, $\phi_{CoM}$ and $\phi_{c_{i}}$ refer to the forward kinematic maps for the center of mass and the $i^{th}$ contact point, respectively.
To ensure kinodynamic feasibility of the resulting trajectories, we incorporate additional contact constraints. Utilizing the contact schedule obtained from previous steps, we impose the following constraints when contact points are active:
\begin{equation}
        \mathbf{f}_{i,k} \in \mathbf{\mathcal{F}} \label{eq:fr_cone}
\end{equation}
\begin{equation}
        c_{i,k} = c_{i,k-1}
        \label{eq:no_slip}
\end{equation}
\begin{equation}
    c_{i,z} = \gamma_{\text{gnd}}(c_{i,x}, c_{i,y})
    \label{eq:contact}
\end{equation}
Equation \eqref{eq:fr_cone} represents the friction cone constraint, where $\mathcal{F}$ is the linear approximation of the friction cone. Equation \eqref{eq:no_slip} enforces a no-slip constraint when a contact point is active in consecutive time steps, whereas Eq. \eqref{eq:contact} ensures contact with the ground. Here, $\gamma_{\text{gnd}}$ represents a height map function that returns the ground height at a specified $(x, y)$ position. Additionally, we enforce a constraint at every timestep to prevent ground penetration by the feet, given by:
\begin{equation}
    c_{i,z} \geq \gamma_{\text{gnd}}(c_{i,x}, c_{i,y}). \label{eq:contact_height}
\end{equation}
To address the numerous instances of self-collisions present in the MoCap data, we use convex approximations of selected links that are prone to collision and impose the following distance constraint:
\begin{equation}
\mathcal{C}_{ij}(\mathbf{q}_k) \geq r_i + r_j \quad \forall (i,j) \in P \label{eq:capsul}
\end{equation}
Where $\mathcal{C}_{ij}$ is a capsule distance function representing the minimum distance between two capsules representing the $i$th and $j$th links. $P$ represents the set of relevant link pairs, and $r_i$ and $r_j$ are the radii of the capsules for links $i$ and $j$ respectively.  Moreover, we impose joint velocity and torque limits in Eqs.\eqref{eq:pos_lim}-\eqref{eq:tau_lim}. Here we approximate the necessary torque as equivalent to that required for producing ground reaction forces and establish limits accordingly \cite{MIT_humanoid, perera2024staccatoe}, which is given by
\begin{equation}
\boldsymbol{\tau}_{j,k} \approx - \mathbf{S}_j^\top \left( \sum_{i=1}^{n_c} \mathbf{J}(\mathbf{q}_k)_{i}^\top \mathbf{f}_{i,k} \right),
\end{equation}
where \(\boldsymbol{\tau}_{j, k} \in \mathbb{R}^{n}\) is the required joint torque at $j$-th joint, \(\mathbf{S}_j \in \mathbb{R}^{n\times n+6}\) is a selection matrix, and \(\mathbf{J}_{i}\in \mathbb{R}^{3\times n+6}\) is the Jacobian of the $i$-th contact point.
\begin{align}
    \mathbf{q}_{min} \le \mathbf{q}_k \le \mathbf{q}_{max} \label{eq:pos_lim}\\
    \mathbf{v}_{min} \le \mathbf{v}_k \le \mathbf{v}_{max}\label{eq:vel_lim}\\
    \boldsymbol{\tau}_{min} \le \boldsymbol{\tau}_{k} \le \boldsymbol{\tau}_{max}\label{eq:tau_lim}
\end{align}
Furthermore, we incorporate the following explicit biomechanics-inspired constraints:
Ankle locking and Approach Angle constraints given by Eqs. \eqref{eq:ankle_lock} and \eqref{eq:approach_angle} respectively.
\begin{equation}
\mathbf{v}_k^{\text{ankle}} = 0, \quad \text{for } k\Delta t \in \left [T^{\text{lock}}, T^{\text{impact}}\right]
\label{eq:ankle_lock}
\end{equation}
\begin{equation}
\theta_{\text{min}} < \cos^{-1}\left(\frac{(\mathbf{p}^{\text{ball}}{xy} - \mathbf{p}^{\text{body}}{xy}) \cdot \hat{\mathbf{x}}}{\|\mathbf{p}^{\text{ball}}{xy} - \mathbf{p}^{\text{body}}{xy}\|}\right) < \theta_{\text{max}}
\label{eq:approach_angle}
\end{equation}
Equation. \eqref{eq:ankle_lock} inspired by \cite{barfield1998biomechanics} ensures that the kicking foot ankle joint velocity $\mathbf{v}_k^{\text{ankle}}$ is zero for before the impact moment, starting from $T^{\text{lock}}_{\text{start}}$ up to the end of the impact period $T^{\text{impact}}$.
Equation \eqref{eq:approach_angle} constrains the approach angle of the robot relative to the ball. It ensures that the angle between the vector from the initial body position $\mathbf{p}^{\text{body}}{xy}$ to the ball position $\mathbf{p}^{\text{ball}}{xy}$ (both projected onto the xy-plane) and the x-axis falls within the range [$\theta_{\text{min}}$, $\theta_{\text{max}}$]. In this work, we set $\theta_{\text{min}} = 24~\si{\degree}$ and $\theta_{\text{max}} = 43~\si{\degree}$, based on biomechanics literature\cite{Egan2007, Shan2011, Shan2022}.
Lastly, we add the dynamics integration constraints.
\begin{align}
\mathbf{q}_{k+1} = \mathbf{q}_k \oplus \mathbf{v}_{k} \Delta t, \\ \label{eq:jt_int}
\mathbf{r}_{k + 1} = \mathbf{r}_k + \mathbf{\dot{r}}_k \Delta t \\
\mathbf{\dot{r}}_{k + 1} = \mathbf{\dot{r}}_k + \mathbf{\ddot{r}}_k \Delta t \\
\mathbf{h}_{k + 1} = \mathbf{h}_k + \mathbf{\dot{h}}_k \Delta t \label{eq:h_int}
\end{align}
The final kinodynamic optimization problem is then given by Eqs. \eqref{eq:cost} to \eqref{eq:ct}.
\begin{align}
    \min_{\substack{\mathbf{r}_k, \dot{\mathbf{r}}_k, \mathbf{h}_k, \mathbf{q}_k, \\ \dot{\mathbf{q}}_k, \mathbf{f_k}, \mathbf{c}_k}} & \quad \sum_{k=1}^{N} J^{\rm ref}_{k}+ J_{k}^{\rm reg}+ J_{k}^{\rm imp} \label{eq:cost}\\ 
    \text{s.t.} \nonumber \\
& \text{Integration constraints (Eqs.  \eqref{eq:jt_int} - \eqref{eq:h_int}})\\
& \text{Centroidal dyn. constraints (Eqs. \eqref{eq:cen_lindyn}-\eqref{eq:cen_angdyn}})\\ 
& \text{Consistency constraints (Eqs. \eqref{eq:ang_consistency}-(\ref{eq:contact_loc}}))\\ 
& \text{Contact constraints (Eqs. \eqref{eq:fr_cone}-\eqref{eq:contact_height}}) \\
& \text{Biomechanics constraints (Eqs. \eqref{eq:ankle_lock}-\eqref{eq:approach_angle})}\\
& \text{Collision constraint (Eqs. \eqref{eq:capsul})}\\
& \text{Joint Limit constraints (Eqs. \eqref{eq:pos_lim}-\eqref{eq:tau_lim})}\label{eq:ct}  
\end{align}
    \label{control}
This optimization problem involves considerable non-linear constraints, although not as much as fullbody dynamics based trajectory optimization, so can be difficult to solve efficiently.  We address this challenge using a two-stage approach: First, we perform a kinematics-only optimization stage where we ignore the centroidal dynamics, and consistency constraints and obtain a purely kinematic trajectory. This stage is equivalent to the optimization-based inverse kinematics approach described in \cite{tang2023humanmimic} and illustrated in Fig.~\ref{fig:offline-framework}(b). We then use the result from the first stage as a warm start for the full optimization problem, which includes both kinematics and centroidal dynamics. Finally, we use the joint position $\mathbf{q}$, velocity $\mathbf{v}$, and ground reaction force $\mathbf{f}$ from the previous step to compute the required torque and verify that torque limits are not violated when full-body dynamics is considered. We do this by first obtaining the joint acceleration $\ddot{\mathbf{q}}$ trajectory through finite differencing of the joint velocity $\mathbf{v}$, and then applying inverse dynamics:
\begin{equation}
\boldsymbol{\tau}^{ID} = \text{ID}(\mathbf{q}, \mathbf{v}, \ddot{\mathbf{q}}, \mathbf{f})
\label{eq:id}
\end{equation}
If necessary, we iterate between the kinodynamics optimization and torque verification steps until all constraints are satisfied. 
\begin{figure}
        \centering
        \includegraphics[width=\linewidth]{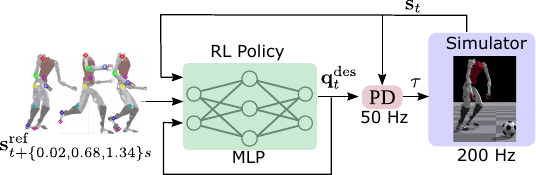}
        \caption{Imitation learning framework}
        \label{fig:control_framework}
\end{figure}
\subsection{Imitation Learning}
Once the physically consistent trajectory is obtained, we use the Proximal Policy Optimization (PPO) algorithm \cite{schulman2017proximal} to train a Reinforcement Learning (RL) control policy in a simulated environment in Isaac Gym \cite{makoviychuk2021isaac} that includes a standard sized soccer ball. This policy has an imitation objective to robustly track the reference trajectory, while making necessary adjustments to account for the mismatch between the centroidal dynamics used in trajectory optimization and the full-body dynamics of the simulation. The RL policy is trained to maximize the following tracking rewards:
\begin{equation}
r_k^{\text{imitation}} =r_k^{\text{keypoint}} + r_k^{\text{joint}} + r_k^{\text{CoM}},
\end{equation}
where 
\begin{equation}
r_k^{\text{keypoint}} = w_k e^{k_k \sum_{m=1}^{M}\| \phi_{FK}^{m}(\mathbf{q}_k^{\rm ref}) - \phi_{FK}^{m}(\mathbf{q}_k) \|_2},
\end{equation}
\begin{equation}
r_k^{\text{joint}} = w_q e^{k_q \| \mathbf{q}_{k}^{\rm ref} - \mathbf{q}_{k} \|_2} + w_v e^{k_v \| \mathbf{v}_{k}^{\rm ref} - \mathbf{v}_{k} \|_2},
\end{equation}
\begin{equation}
r_k^{\text{CoM}} = w_c e^{k_c \| \mathbf{p}_{comk}^{\rm ref} -\mathbf{p}_{comk} \|_2}. 
\end{equation}
Here, $w_k$, $w_q$, $w_v$, and $w_c$ are weight parameters for each reward component
$k_k$, $k_q$, $k_v$, and $k_c$ are scaling factors. $\phi_{FK}^{m}(\cdot)$ is the forward kinematics map for the $m$-th keypoint.
In addition to the imitation objective, we incorporate the reward to maximize the ball's velocity toward the desired direction: 
\begin{equation}
r_k^{\text{ball}} = w_{\text{ball}} * (e^{\max(0, \mathbf{v}^{\text{ball}} \cdot \mathbf{n}^{\text{target}})} - 1) \label{eq:impact_rew}
\end{equation}
where $\mathbf{v}^{\text{ball}}$ is the velocity vector of the ball and $\mathbf{n}^{\text{target}}$ is the unit vector representing the desired direction of ball travel.
\subsection{Sample Efficient RL Training Through Reference Guided Early Termination Strategy}
Previous works in the graphics community have demonstrated the importance of early termination for sample-efficient training of motion imitation \cite{peng2018deepmimic, luo2023perpetual}. Early termination prevents unnecessary local optima behaviors by assigning zero reward after the agent enters undesirable states, thereby avoiding unnecessary exploration and improving sample efficiency.
Luo et al. \cite{luo2023perpetual} used a Reference-based Early Termination (RET)  strategy, where the agent is terminated if the average link position of the robot deviates from the reference motion. However, this kind of termination strategy based on motion references without consideration of the robot's dynamics faces challenges in yielding the desired behavior, especially for highly dynamic motions like a soccer kick. This is because faithfully following the kinematics reference alone does not guarantee that physical limits, such as torque limits, are not violated. Therefore, the agent might need to deviate from the reference significantly to accommodate the limits not considered in the motion planning stage.

This challenge is a primary reason why we spent significant efforts to create a kinodynamic trajectory optimization framework. The obtained optimal trajectories are guaranteed the system to operate within the physical limits; thereby the trajectories can serve as a trustful prior for constraining the search space during RL training through early termination. In particular, we use the following distance threshold-based RET strategy to achieve sample-efficient exploration:
\begin{equation}
\text{Terminate if } \max_{i \in \mathcal{L}} \|\phi_{FK}^i(\mathbf{q}) - \phi_{FK}^i(\mathbf{q}^{\text{ref}})\|_2 > \Bar{d}
\end{equation}
That is, terminate the episode if the position of any link of the robot in the link set $\mathcal{L}$ deviates from the reference by a distance greater than the threshold $\Bar{d}$.

\begin{figure}
    \centering
    \includegraphics[width=0.8\linewidth]{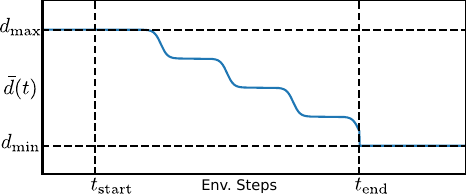}    \caption{Stair Case Distance Threshold Function}
    \label{fig:dt}
\end{figure}

We consider two kinds of thresholds for our termination strategy. The first is a fixed threshold, where a single distance threshold is used throughout the training process. The second one a time-varying threshold, where the threshold is updated as the training progresses, given by Eq. \eqref{eq:time_varying_threshold}. Specifically, we continuously reduce the distance threshold over time following a staircase-like function, as depicted in Fig. \ref{fig:dt}, as the policy becomes better at tracking the reference. This approach allows for finer and more precise tracking, enabling the policy to maintain biomechanical characteristics captured in the human MoCap data while allowing for necessary adaptations to the robot's specific dynamics.
\begin{equation}
\Bar{d}(t) = \begin{cases} 
    d_{\text{max}} &\text{if } t < t_{\text{start}} \\
    d_{\text{min}} &\text{if } t \geq t_{\text{end}} \\
    d_{\text{max}} - \sum_{i=1}^{n} \Delta h \cdot \sigma\left(\gamma \cdot \frac{t - (t_{\text{start}} + i\Delta t)}{\Delta t}\right) & \text{otherwise}
\label{eq:time_varying_threshold}
\end{cases}
\end{equation}
where $\sigma$ is the sigmoid function given by $\sigma(x) = \frac{1}{1 + e^{-x}}$, $\gamma$ is a smoothness parameter, $\Delta d = \frac{d_{\text{max}} - d_{\text{min}}}{n}$ and 
$\Delta t = \frac{t_{\text{end}}-t_{\text{start}}}{n}$.
\section{Experimental Setup}
\subsection{Experimental Platform}
\begin{table}
\small
\centering
\caption{Prestoe joint torque and velocity limits}
\label{tab:joint_limits}
\begin{tabular}{|c|c|c|c|}
\hline
\textbf{Group} & \textbf{Joint} & \textbf{Max Torque} & \textbf{Max Speed} \\
\hline
Leg & Hip  & 48 N·m & 20 rad/s \\
& Knee & 200 N·m & 10 rad/s \\
& Ankle & 100 N·m & 10 rad/s \\
& Toe & 10 N·m & 20 rad/s \\
\hline
Arm & Shoulder &  &  \\
& Elbow & 18 N·m & 40 rad/s \\
& Wrist  & &  \\
\hline
Torso & Torso  & 18 N·m & 40 rad/s \\
\hline
\end{tabular}%
\end{table}
In this study, we use PresToe, illustrated in Fig.~\ref{fig:prestoe}, for all our simulation experiments. PresToe is a 25 DoF humanoid robot, comprised of two 7 DoF legs, each with a distinct 1 DoF toe actuation, two 5 DoF arms, and a 1 DoF torso. The robot stands at a height of 1.3 meters in its nominal configuration and weighs 29.5 kg. The joint torque and velocity limits are listed in Table \ref{tab:joint_limits}. PresToe is an ideal candidate for replicating human-like movements, given its comparable size and anthropomorphic design. The leg design is based on our earlier work on Staccatoe \cite{perera2024staccatoe} and has a unique co-actuation coupling mechanism that allows the robot to generate large knee and ankle torques of up to 200 N·m for the knee and 100 N·m for the ankle, depending on the robot's joint configuration.

\subsection{Reference Motion Generation and RL Policy Training}
 We generate reference motion by kinodynamically retargeting MoCap data of a human performing a soccer kick, obtained from the CMU Mocap Dataset \cite{cmu_mocap}. The original motion is 3.7 seconds long, recorded at 120 Hz. We subsample it to 30 Hz, resulting in a trajectory optimization horizon of 112 time steps. We follow the four steps described in Section~\ref{sec:kto} to perform the retargeting.

The overall Imitation Learning framework is depicted in Fig.~\ref{fig:control_framework}. As the figure shows, the observation space of the policy is given by 
\begin{equation}
 \mathbf{o}_t = \left[\mathbf{s}_{t+\{0.02, 0.68, 1.34\}s}^{\text{ref}}, \mathbf{s}_t, \mathbf{q}^{\text{des}}_{t-0.02s}\right],   
\end{equation}
where $\mathbf{s}_{t+{0.02, 0.68, 1.34}s}^{\text{ref}}$ represents the future desired reference states at 0.02s, 0.68s, and 1.34s ahead of the current time, $\mathbf{s}_t$ is the current state represented by ${\mathbf{q}, \mathbf{v}}$, and $\mathbf{q}^{\text{des}}_{t-0.02}$ is the action at the previous time step which is a desired joint position command. We convert this position command to a torque command using the following PD control law:
\begin{equation}
\boldsymbol{\tau} = \mathbf{k}_p(\mathbf{q}^{des}-\mathbf{q}) - \mathbf{k}_d \mathbf{v},
\end{equation}
where $\mathbf{k}_p$ and $\mathbf{k}_d$ are PD gains. 
In all experiments, the episode begins with the robot initialized with a random state sampled from the reference motion similar to \cite{peng2018deepmimic}. For both the policy and value networks, we use a multi-layer perceptron neural network consisting of two hidden layers with sizes 1024 and 512, respectively, alongside ReLU activation.

\subsection{Sample Efficiency Experiment}
We assess the effectiveness of introducing dynamics in the offline motion planning stage by comparing motion tracking policies trained with solely imitation objectives using three different reference types. The first is a kinodynamic reference trajectory that takes into consideration the centroidal dynamics of the robot and imposes appropriate torque limits. The second is a purely kinematics-based motion reference similar to \cite{tang2023humanmimic}, which retains artifacts such as foot skating present in the reference MoCap data. The third is another purely kinematic reference but that does not involve artifacts like foot sliding. This reference is obtained by leveraging the same contact schedule used to obtain the kinodynamic trajectory.
\subsection{Ball Kicking Experiment}
In this experiment, we first train a motion tracking policy solely with the motion imitation objective using a time-varying distance threshold approach in Eq.~\eqref{eq:time_varying_threshold}. The threshold changes from 30 cm to 10 cm over 1 billion training steps. The goal here is to track the reference as closely as possible and retain biomechanical insights captured in the human MoCap data.
We then adapt this policy for a high-impact soccer ball kicking task. To achieve this, we introduce a standard FIFA size 5 ball (weighing 430g with a diameter of 22 cm) into the simulation. We further train the policy to maximize ball velocity in the forward direction using the impact reward in Eq. \eqref{eq:impact_rew}, in addition to the imitation objective. Note that in this work, we assume a fixed ball position and we do not provide it as an observation to the policy. Instead, it is inferred indirectly through the ball impact reward.
\section{Results and Discussion}
\begin{figure*}[!ht]
    \centering
    \includegraphics[width=\linewidth]{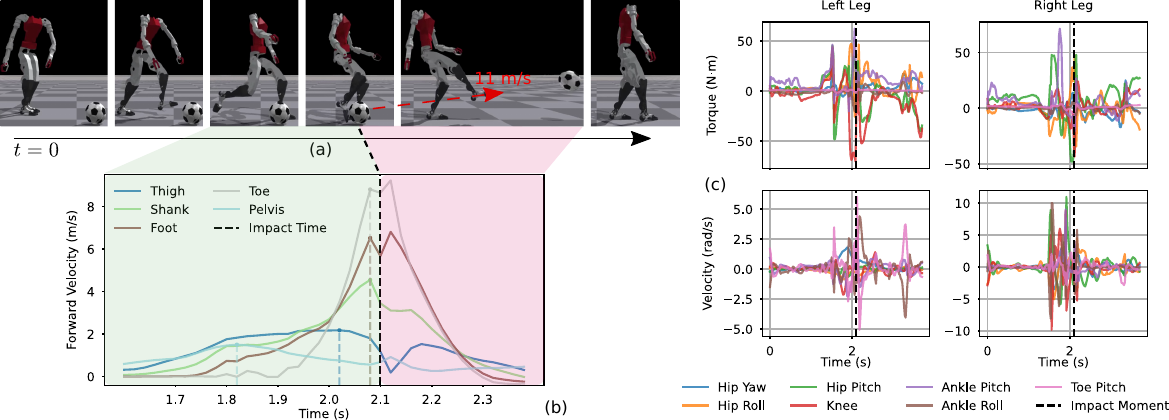}
\caption{Human-like instep soccer kicking with Prestoe: (a) Motion timelapse from left to right: the robot executes the classic instep kick phases, i.e., momentum-building step, windup, kicking leg acceleration, impact, and follow-through. (b) Forward velocity of the kicking leg links moments before impact and during follow-through phase; vertical dashed lines indicate the moment each link reaches its maximum speed. (c) Torque and velocity profiles throughout the motion for both kicking and non-kicking legs.}
    \label{fig:kick-motion}
\end{figure*}

\begin{figure}
    \centering
    \includegraphics[width=\linewidth]{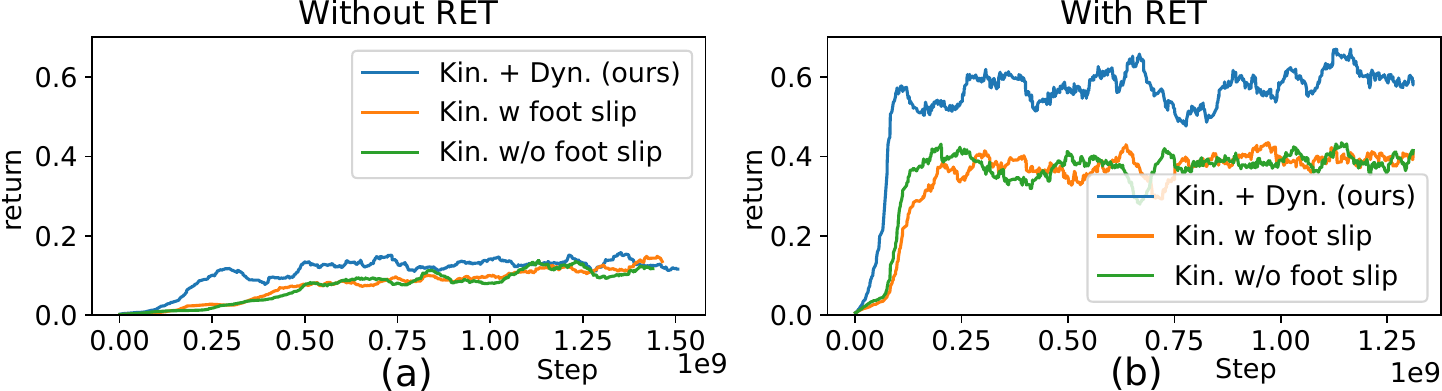}
    \caption{Training progress measured by normalized imitation reward return vs. environment steps for three reference types: Kinematic-only with foot slippage artifacts (Kin w/ foot slip), Kinematic-only without foot slippage (Kin w/o foot slip), and Kinodynamic reference. (a) Without reference-based early termination (RET). (b) With reference-based early termination.}
    \label{fig:training}
\end{figure}

Fig.~\ref{fig:training} illustrates the imitation reward return as training progresses under two scenarios. Fig.~\ref{fig:training} (a) shows the performance without reference-based early termination (RET), while Fig~\ref{fig:training}  (b) demonstrates the results with RET using a distance threshold of 30 cm. Both Figs \ref{fig:training} (a) and (b) display the training performance when three different reference types are used: kinematic-only with motion artifact like foot sliding, kinematic-only but without motion artifacts, and kinodynamic references.
As evident from the figure, the policy trained with kinodynamically consistent trajectories learns faster and achieves higher returns in both scenarios. When RET is deployed, there is a noticeable advantage, especially when motion artifacts like foot slippage are removed. The policy tends to learn faster in this case, although not as effectively as with references that consider dynamics.

Table~\ref{tab:pos_error} presents the reference tracking performance of the trained policies, measured using the average link position error between the policy and the reference at each time step. The table clearly demonstrates the importance of RET. In fact, as can be observed in the attached video, without RET, the policies fail to learn to emulate the motion with just reward signals alone in one billion steps for all three reference types. However, with when RET is used, all three policies successfully learned to imitate the motion, with the policy trained on kinodynamically consistent trajectories learning significantly faster and more closely resembling the reference.
The kinodynamic reference's superior performance can be attributed to its consideration of both kinematics and dynamics of the robot as well as its physical limits, providing a more realistic and achievable trajectory for the policy to imitate. 

Fig.~\ref{fig:kick-motion}(a) presents a snapshot sequence of the robot performing an instep kick using the kicking policy. 
The kick follows the six steps of an instep kick: approach, support phase, windup, leg swing, ball contact, and follow-through phases, 
achieving ball speeds exceeding $11~\si{\meter\per\second}$. This speed is approximately double the $6.6~\si{\meter\per\second}$ 
instep kick achieved by the ground-anchored single-leg robot detailed in \cite{Haddadin2009}, 
and about 40\% of the speed an average human player can achieve \cite{radja2019ball}.
This performance is noteworthy considering the robot's size and limited torque capability. The robot has a maximum hip pitch torque of $48 \text{N.m}$, compared to the average human player's hip joint flexion/extension capability of over $250 \text{N.m}$ \cite{sakamoto2016kinetic}.
Fig.~\ref{fig:kick-motion}(b) showcases the velocities of the kicking leg's links in the forward direction before and after ball impact. We observe that the pelvis reaches its maximum speed and starts to decelerate before the thigh reaches its maximum speed and decelerates. Moreover, the shank, foot, and toe links reach their maximum speeds at the moment of impact. This pattern exhibits the classic proximal-to-distal kinetic energy transfer present in optimal soccer kicks, as discussed in biomechanics literature \cite{Lees1998, Lees2010, Shan2011, Shan2022}.
Finally, Fig.~\ref{fig:kick-motion}(c) shows the torque and velocity profiles of the kicking (Right leg) and non-kicking leg joints throughout the entire motion and demonstrates that they remain within the physical limits of the robot.
\begin{table}
\small
\centering
\caption{Average Link Position Tracking Error}
\label{tab:pos_error}
\begin{tabular}{|c|c|c|c|}
\hline
\textbf{Ref. Type} & \textbf{Without RET} & \textbf{With RET}\\
\hline
Kin. w foot slip & 0.66 +/- 0.47 m  & 0.14 +/- 0.03 m  \\
\hline
Kin. w/o foot slip& 0.61 +/- 0.51 m & 0.114 +/- 0.06 m \\
\hline
Kin. + Dyn. (ours) & \textbf{0.60 +/- 0.50} m & \textbf{0.09 +/- 0.02} m\\
\hline
\end{tabular}%
\end{table}

\section{Conclusion and Future work}
In this study, we developed a motion planning and imitation learning-based control framework informed by human biomechanics to facilitate dynamic and powerful soccer kicks with a humanoid robot. This approach significantly surpasses the limited kick-power outputs observed in previous works which used quasi-static and walk-kick type kicking techniques.  Additionally, we demonstrated the effectiveness of kinodynamic motion retargeting, which produces physically consistent trajectories for efficiently learning control policies, as opposed to kinematics-only approaches that don’t take into account the dynamics of the robot.
In future work, we aim to extend and validate our approach with other dynamic motions beyond soccer kicks. Additionally, as all experiments in this study were conducted in a simulated environment, we plan to validate these trajectories on a physical robot. 
\section*{Acknowledgment}
This material is based upon work supported by the National Science Foundation under Grant No. 2220924. 
\bibliographystyle{IEEEtran}
\bibliography{references}

\begin{thebibliography}{10}
\providecommand{\url}[1]{#1}
\csname url@samestyle\endcsname
\providecommand{\newblock}{\relax}
\providecommand{\bibinfo}[2]{#2}
\providecommand{\BIBentrySTDinterwordspacing}{\spaceskip=0pt\relax}
\providecommand{\BIBentryALTinterwordstretchfactor}{4}
\providecommand{\BIBentryALTinterwordspacing}{\spaceskip=\fontdimen2\font plus
\BIBentryALTinterwordstretchfactor\fontdimen3\font minus \fontdimen4\font\relax}
\providecommand{\BIBforeignlanguage}[2]{{%
\expandafter\ifx\csname l@#1\endcsname\relax
\typeout{** WARNING: IEEEtran.bst: No hyphenation pattern has been}%
\typeout{** loaded for the language `#1'. Using the pattern for}%
\typeout{** the default language instead.}%
\else
\language=\csname l@#1\endcsname
\fi
#2}}
\providecommand{\BIBdecl}{\relax}
\BIBdecl

\bibitem{Kitano1998}
\BIBentryALTinterwordspacing
H.~Kitano and M.~Asada, ``The {RoboCup} humanoid challenge as the millennium challenge for advanced robotics,'' \emph{Advanced Robotics}, vol.~13, no.~8, pp. 723--736, Jan. 1998. [Online]. Available: \url{https://doi.org/10.1163/156855300x00061}
\BIBentrySTDinterwordspacing

\bibitem{Gerndt2015}
\BIBentryALTinterwordspacing
R.~Gerndt, D.~Seifert, J.~H. Baltes, S.~Sadeghnejad, and S.~Behnke, ``Humanoid robots in soccer: Robots versus humans in {RoboCup} 2050,'' \emph{{IEEE} Robotics Automation Magazine}, vol.~22, no.~3, pp. 147--154, Sep. 2015. [Online]. Available: \url{https://doi.org/10.1109/mra.2015.2448811}
\BIBentrySTDinterwordspacing

\bibitem{cmu}
H.~Schempf, C.~Kraeuter, and M.~Blackwell, ``Roboleg: a robotic soccer-ball kicking leg,'' in \emph{Proceedings of 1995 IEEE International Conference on Robotics and Automation}, vol.~2, 1995, pp. 1314--1318 vol.2.

\bibitem{Haddadin2009}
\BIBentryALTinterwordspacing
S.~Haddadin, T.~Laue, U.~Frese, S.~Wolf, A.~Albu-Sch\"{a}ffer, and G.~Hirzinger, ``Kick it with elasticity: Safety and performance in human{\textendash}robot soccer,'' \emph{Robotics and Autonomous Systems}, vol.~57, no.~8, pp. 761--775, Jul. 2009. [Online]. Available: \url{https://doi.org/10.1016/j.robot.2009.03.004}
\BIBentrySTDinterwordspacing

\bibitem{Vahidi2015}
\BIBentryALTinterwordspacing
M.~Vahidi and S.~A.~A. Moosavian, ``Dynamics of a 9-{DoF} robotic leg for a football simulator,'' in \emph{2015 3rd {RSI} International Conference on Robotics and Mechatronics ({ICROM})}.\hskip 1em plus 0.5em minus 0.4em\relax {IEEE}, Oct. 2015. [Online]. Available: \url{https://doi.org/10.1109/icrom.2015.7367803}
\BIBentrySTDinterwordspacing

\bibitem{nao}
J.~Müller, T.~Laue, and T.~Röfer, ``Kicking a ball – modeling complex dynamic motions for humanoid robots,'' 01 2010, pp. 109--120.

\bibitem{barrett2010controlled}
S.~Barrett, K.~Genter, T.~Hester, M.~Quinlan, and P.~Stone, ``Controlled kicking under uncertainty,'' in \emph{The Fifth Workshop on Humanoid Soccer Robots at Humanoids}, 2010.

\bibitem{Hester2010}
\BIBentryALTinterwordspacing
T.~Hester, M.~Quinlan, and P.~Stone, ``Generalized model learning for reinforcement learning on a humanoid robot,'' in \emph{2010 {IEEE} International Conference on Robotics and Automation}.\hskip 1em plus 0.5em minus 0.4em\relax {IEEE}, May 2010. [Online]. Available: \url{https://doi.org/10.1109/robot.2010.5509181}
\BIBentrySTDinterwordspacing

\bibitem{Leottau2015}
L.~Leottau, C.~Celemin, and J.~Ruiz-del Solar, ``Ball dribbling for humanoid biped robots: A reinforcement learning and fuzzy control approach,'' in \emph{RoboCup 2014: Robot World Cup XVIII}, R.~A.~C. Bianchi, H.~L. Akin, S.~Ramamoorthy, and K.~Sugiura, Eds.\hskip 1em plus 0.5em minus 0.4em\relax Cham: Springer International Publishing, 2015, pp. 549--561.

\bibitem{sung2011}
\BIBentryALTinterwordspacing
D.~Closson, Ed., \emph{Tectonics}.\hskip 1em plus 0.5em minus 0.4em\relax {InTech}, Feb. 2011. [Online]. Available: \url{https://doi.org/10.5772/567}
\BIBentrySTDinterwordspacing

\bibitem{Teixeira2020}
\BIBentryALTinterwordspacing
H.~Teixeira, T.~Silva, M.~Abreu, and L.~P. Reis, ``Humanoid robot kick in motion ability for playing robotic soccer,'' in \emph{2020 {IEEE} International Conference on Autonomous Robot Systems and Competitions ({ICARSC})}.\hskip 1em plus 0.5em minus 0.4em\relax {IEEE}, Apr. 2020. [Online]. Available: \url{https://doi.org/10.1109/icarsc49921.2020.9096073}
\BIBentrySTDinterwordspacing

\bibitem{Rezaeipanah2020}
\BIBentryALTinterwordspacing
A.~Rezaeipanah, P.~Amiri, and S.~Jafari, ``Performing the kick during walking for {RoboCup} 3d soccer simulation league using reinforcement learning algorithm,'' \emph{International Journal of Social Robotics}, vol.~13, no.~6, pp. 1235--1252, Nov. 2020. [Online]. Available: \url{https://doi.org/10.1007/s12369-020-00712-2}
\BIBentrySTDinterwordspacing

\bibitem{asimo_demo_2014}
\BIBentryALTinterwordspacing
A.~Express, ``Honda's asimo: the penalty-taking, bar-tending robot,'' 2014, youTube video. [Online]. Available: \url{https://www.youtube.com/watch?v=QdQL11uWWcI}
\BIBentrySTDinterwordspacing

\bibitem{Shan2011}
\BIBentryALTinterwordspacing
G.~Shan and X.~Zhang, ``From 2d leg kinematics to 3d full-body biomechanics-the past, present and future of scientific analysis of maximal instep kick in soccer,'' \emph{Sports Medicine, Arthroscopy, Rehabilitation, Therapy Technology}, vol.~3, no.~1, Oct. 2011. [Online]. Available: \url{https://doi.org/10.1186/1758-2555-3-23}
\BIBentrySTDinterwordspacing

\bibitem{bio_youtube_video}
BiomechanicsMMU, ``Football free kick - slow motion video,'' \url{https://www.youtube.com/watch?v=lBMA2wWuqh8}, 2009, accessed: Wed Oct 25 22:32:03 EDT 2023.

\bibitem{Shan2022}
\BIBentryALTinterwordspacing
G.~Shan, ``?{\textemdash}a state-of-the-art review,'' \emph{Applied Sciences}, vol.~12, no.~21, p. 10886, Oct. 2022. [Online]. Available: \url{https://doi.org/10.3390/app122110886}
\BIBentrySTDinterwordspacing

\bibitem{Egan2007}
\BIBentryALTinterwordspacing
C.~D. Egan, M.~H.~G. Verheul, and G.~J.~P. Savelsbergh, ``Effects of experience on the coordination of internally and externally timed soccer kicks,'' \emph{Journal of Motor Behavior}, vol.~39, no.~5, pp. 423--432, Sep. 2007. [Online]. Available: \url{https://doi.org/10.3200/jmbr.39.5.423-432}
\BIBentrySTDinterwordspacing

\bibitem{barfield1998biomechanics}
W.~R. Barfield, ``The biomechanics of kicking in soccer,'' \emph{Clinics in sports medicine}, vol.~17, no.~4, pp. 711--728, 1998.

\bibitem{Lees1998}
\BIBentryALTinterwordspacing
A.~Lees and L.~Nolan, ``The biomechanics of soccer: A review,'' \emph{Journal of Sports Sciences}, vol.~16, no.~3, pp. 211--234, Jan. 1998. [Online]. Available: \url{https://doi.org/10.1080/026404198366740}
\BIBentrySTDinterwordspacing

\bibitem{Kellis2007}
E.~Kellis and A.~Katis, ``{{B}iomechanical characteristics and determinants of instep soccer kick},'' \emph{J Sports Sci Med}, vol.~6, no.~2, pp. 154--165, 2007.

\bibitem{Brophy2007}
\BIBentryALTinterwordspacing
R.~H. Brophy, S.~I. Backus, B.~S. Pansy, S.~Lyman, and R.~J. Williams, ``Lower extremity muscle activation and alignment during the soccer instep and side-foot kicks,'' \emph{Journal of Orthopaedic Sports Physical Therapy}, vol.~37, no.~5, pp. 260--268, May 2007. [Online]. Available: \url{https://doi.org/10.2519/jospt.2007.2255}
\BIBentrySTDinterwordspacing

\bibitem{Lees2010}
\BIBentryALTinterwordspacing
A.~Lees, T.~Asai, T.~B. Andersen, H.~Nunome, and T.~Sterzing, ``The biomechanics of kicking in soccer: A review,'' \emph{Journal of Sports Sciences}, vol.~28, no.~8, pp. 805--817, Jun. 2010. [Online]. Available: \url{https://doi.org/10.1080/02640414.2010.481305}
\BIBentrySTDinterwordspacing

\bibitem{dai}
H.~Dai, A.~Valenzuela, and R.~Tedrake, ``Whole-body motion planning with centroidal dynamics and full kinematics,'' in \emph{2014 IEEE-RAS International Conference on Humanoid Robots}, 2014, pp. 295--302.

\bibitem{tang2023humanmimic}
A.~Tang, T.~Hiraoka, N.~Hiraoka, F.~Shi, K.~Kawaharazuka, K.~Kojima, K.~Okada, and M.~Inaba, ``Humanmimic: Learning natural locomotion and transitions for humanoid robot via wasserstein adversarial imitation,'' \emph{arXiv preprint arXiv:2309.14225}, 2023.

\bibitem{yan2023imitationnet}
Y.~Yan, E.~V. Mascaro, and D.~Lee, ``Imitationnet: Unsupervised human-to-robot motion retargeting via shared latent space,'' in \emph{2023 IEEE-RAS 22nd International Conference on Humanoid Robots (Humanoids)}.\hskip 1em plus 0.5em minus 0.4em\relax IEEE, 2023, pp. 1--8.

\bibitem{cheng2024expressive}
X.~Cheng, Y.~Ji, J.~Chen, R.~Yang, G.~Yang, and X.~Wang, ``Expressive whole-body control for humanoid robots,'' \emph{arXiv preprint arXiv:2402.16796}, 2024.

\bibitem{he2024learning}
T.~He, Z.~Luo, W.~Xiao, C.~Zhang, K.~Kitani, C.~Liu, and G.~Shi, ``Learning human-to-humanoid real-time whole-body teleoperation,'' \emph{arXiv preprint arXiv:2403.04436}, 2024.

\bibitem{nakaoka2005task}
S.~Nakaoka, A.~Nakazawa, F.~Kanehiro, K.~Kaneko, M.~Morisawa, and K.~Ikeuchi, ``Task model of lower body motion for a biped humanoid robot to imitate human dances,'' in \emph{2005 IEEE/RSJ International Conference on Intelligent Robots and Systems}.\hskip 1em plus 0.5em minus 0.4em\relax IEEE, 2005, pp. 3157--3162.

\bibitem{rempe2020contact}
D.~Rempe, L.~J. Guibas, A.~Hertzmann, B.~Russell, R.~Villegas, and J.~Yang, ``Contact and human dynamics from monocular video,'' in \emph{Computer Vision--ECCV 2020: 16th European Conference, Glasgow, UK, August 23--28, 2020, Proceedings, Part V 16}.\hskip 1em plus 0.5em minus 0.4em\relax Springer, 2020, pp. 71--87.

\bibitem{makoviychuk2021isaac}
V.~Makoviychuk, L.~Wawrzyniak, Y.~Guo, M.~Lu, K.~Storey, M.~Macklin, D.~Hoeller, N.~Rudin, A.~Allshire, A.~Handa \emph{et~al.}, ``Isaac gym: High performance gpu-based physics simulation for robot learning,'' \emph{arXiv preprint arXiv:2108.10470}, 2021.

\bibitem{radja2019ball}
A.~Radja, L.~P. Kuvavci{c, Goran and De Giorgio, Andrea and Sellami, Maha and Ardigo}, N.~L. Bragazzi, and J.~Padulo, ``The ball kicking speed: A new, efficient performance indicator in youth soccer,'' \emph{Plos one}, vol.~14, no.~5, p. e0217101, 2019.

\bibitem{sakamoto2016kinetic}
K.~Sakamoto, N.~Numazu, S.~Hong, and T.~Asai, ``Kinetic analysis of instep and side-foot kick in female and male soccer players,'' \emph{Procedia engineering}, vol. 147, pp. 214--219, 2016.

\bibitem{MIT_humanoid}
\BIBentryALTinterwordspacing
M.~Chignoli, D.~Kim, E.~Stanger{-}Jones, and S.~Kim, ``The {MIT} humanoid robot: Design, motion planning, and control for acrobatic behaviors,'' \emph{CoRR}, vol. abs/2104.09025, 2021. [Online]. Available: \url{https://arxiv.org/abs/2104.09025}
\BIBentrySTDinterwordspacing

\bibitem{perera2024staccatoe}
N.~Perera, S.~Yu, D.~Marew, M.~Tang, K.~Suzuki, A.~McCormack, S.~Zhu, Y.-J. Kim, and D.~Kim, ``Staccatoe: A single-leg robot that mimics the human leg and toe,'' \emph{arXiv preprint arXiv:2404.05039}, 2024.

\bibitem{schulman2017proximal}
J.~Schulman, F.~Wolski, P.~Dhariwal, A.~Radford, and O.~Klimov, ``Proximal policy optimization algorithms,'' \emph{arXiv preprint arXiv:1707.06347}, 2017.

\bibitem{peng2018deepmimic}
X.~B. Peng, P.~Abbeel, S.~Levine, and M.~Van~de Panne, ``Deepmimic: Example-guided deep reinforcement learning of physics-based character skills,'' \emph{ACM Transactions On Graphics (TOG)}, vol.~37, no.~4, pp. 1--14, 2018.

\bibitem{luo2023perpetual}
Z.~Luo, J.~Cao, K.~Kitani, W.~Xu \emph{et~al.}, ``Perpetual humanoid control for real-time simulated avatars,'' in \emph{Proceedings of the IEEE/CVF International Conference on Computer Vision}, 2023, pp. 10\,895--10\,904.

\bibitem{cmu_mocap}
F.~De~la Torre, J.~Hodgins, A.~Bargteil, X.~Martin, J.~Macey, A.~Collado, and P.~Beltran, ``Guide to the carnegie mellon university multimodal activity (cmu-mmac) database,'' 2009.

\end{thebibliography}

\end{document}